\documentclass[runningheads]{llncs}
\usepackage{graphicx}
\usepackage{upgreek}
\usepackage{xcolor}
\usepackage[linesnumbered,ruled,vlined]{algorithm2e}
\usepackage{tabularx}
\usepackage{amsmath,bm}
\usepackage[backref]{hyperref} 

\newcolumntype{Y}{>{\centering\arraybackslash}X}

\SetKwInput{KwInput}{Input}                
\SetKwInput{KwOutput}{Output}
\begin{document}
%
\title{Hybrid Function Sparse Representation towards Image Super Resolution}
%
%
\author{Junyi Bian\inst{1}
\and Baojun Lin\inst{1,2}
\and Ke Zhang\inst{1}}
\authorrunning{F. Author et al.}
%
\institute{
  ShanghaiTech University, Shanghai, China \\
  \email{bianjy@shanghaitech.edu.cn}\\
  \and
  Chinese Academy of Sciences, Shanghai, China \\
  \email{linbaojun@aoe.ac.cn}
}
\maketitle              
\begin{abstract}

Sparse representation with training-based dictionary has been shown successful on super resolution(SR)
but still have some limitations.
Based on the idea of making the magnification of function curve without losing its fidelity,
 we proposed a function based dictionary on sparse representation for super resolution,
 called hybrid function sparse representation (HFSR).
The dictionary we designed is directly generated by preset hybrid functions without additional training,
 which can be scaled to any size as is required due to its scalable property.
We mixed approximated Heaviside function (AHF), sine function and DCT function as the dictionary.
Multi-scale refinement is then proposed to utilize the scalable property of the dictionary to improve the results.
 In addition, a reconstruct strategy is adopted to deal with the overlaps.
The experiments on ’Set14’ SR dataset show that our method has an excellent performance  particularly
with regards to images containing rich details and contexts compared with non-learning based state-of-the art methods.
\end{abstract}

\section{Introduction}

The target of the single image super resolution (SISR) is to produce a high resolution (HR) image from an observed low resolution (LR) image.
 This is meaningful due to its valuable applications in many real-world scenarios such as remoting sensing, magnetic resonance and surveillance camera.
 However, SR is an ill-posed issue which is very difficult to obtain the optimal solution.
The approaches for SISR can be divided into three categories including
interpolation-based methods \cite{zhang_2006,deng_2016,getreuer_2011,li_2001,dai_2007},
reconstruction-based methods \cite{babacan_2011,sun_2008,dong_2013,dong_2011,fattal_2007},
and learning-based methods\cite{freeman_2000,glasner_2009,yang_2010,dong_2014,sajjadi_2017,ledig_2017}.
Reconstruction-based methods and learning-based methods often yield more accurate results than interpolation-based algorithm,
 since those methods can acquire more information from statistical prior and external data
even though more time-consuming.
Algorithm can be integrated from multiple categories mentioned above, which is not strictly independent.

As classic approaches in SR, interpolation-based methods
follow a basic strategy to construct the unknown pixel of HR image such as bicubic, bilinear and nearest-neighbor(NN).
However, nearest-neighbor interpolation always causes a jag effect.
 In this aspect, bicubic and bilinear perform better than NN and are widely used as zooming algorithms
 but easier to generate a blur effect on images.

Reconstruction-based methods usually model an objective function with reconstruction constraints from image priors and
regularization terms to lead better solutions for the inverse problem. The image priors for
reconstructions include the gradient priors \cite{sun_2008}, sparsity priors \cite{dong_2013,dong_2011,kim_2010}
and edge priors \cite{fattal_2007}.
Those methods usually restore the images with sharp edges and less noises, but will be invalid when the upscaling
factor becomes larger.

Learning-based algorithms mainly explore the relationship between LR and HR image patches.
In the early time, Freeman et al. \cite{freeman_2000} predict the HR image from LR input by Markov Random Field trained by
belief propagation.
Sparse-representation for SISR is firstly proposed by Yang et al.\cite{yang_2010} and achieves state-of-the-art performance.
Zhang et al. \cite{zhang_2016} propose a novel classification method based on sparse coding autoextractor.
Zeyde et al. \cite{zeyde_2010} improve the algorithm by using K-SVD \cite{ksvd_2006} based dictionary and
orthogonal matching pursuit (OMP) optimization method.
Recently, deep neural network has been extensively used in super-resolution task.
Dong et al.  \cite{dong_2014} firstly design an end-to-end mapping from LR to HR called SRCNN.
Generative adversarial network \cite{gan_2014} is introduced to SR by \cite{sajjadi_2017,ledig_2017}
to encourage the network to favor solutions that look more like natural images.
The learning-based algorithms are not generally promoted, since most of the algorithms only focus on
the specific upscaling factor and need to be retrained on external large dataset once the upscaling
 factor is changed.

In this paper, we propose hybrid function
sparse representation (HFSR), which uses function-based dictionary to replace conventional training-based dictionary.
The special property of such dictionary are utilized to reinforce the SR algorithm.
Primary idea of the proposed method is that the upscaling of the function curve will keep it's fidelity,
and to represent the patches from LR image by linear combination of functions.
We select approximate Heaviside function (AHF),
discrete cosine transformation (DCT) function, and sine function to form the dictionary.
Unlike the work of \cite{deng_2016}, we adopt diverse functions instead of single AHF, and
do not mix the sparsity and intensity priors together for representation which needs ADMM to solve the objective loss.
As a consequence, approach applying with sparse representation performs much faster than the algorithm in \cite{deng_2016}.
Different from the conventional sparse representation method,
our algorithm requires no additional training for dictionary on high resolution images,
and there is no scale constraint since the dictionary can be scaled up to any size.
Once being represented by hybrid functions,
dictionary can be stored by parameters rather than matrixs with large size.
The algorithm we design is related to the interpolation-based method
and reconstruction-based method.
In view of this, we compare our results with two
 interpolation-based algorithms and another reconstruction-based algorithm \cite{glasner_2009}.


The remaining part of the paper is organized as follows:
Section \ref{2} briefly introduces the sparse representation
algorithm for single image super resolution.
Section \ref{3} describes the proposed HFSR approach
from dictionary designing to multi-scale refinement and then to the construction of whole image.
Section \ref{4} discusses the results of the proposed method and
 the paper is concluded in Section \ref{5}.

\section{Super Resolution via sparse Representation Preliminaries}  \label{2}

The goal of single image super resolution is to recover the HR image $\mathbf{x}$ by giving an observed LR image $\mathbf{y}$.
Relationship between LR images and HR images is denoted as:
  \begin{equation} \label{eq1}
    \mathbf{y} = D \mathbf{x} + v
  \end{equation}
where $D$ is the degradation matrix combining with down sampling operator and blur operator, and $v$ is the independent Gaussian noise.
Sparse representation method processes the image in patch level, which means a LR image is divided into several small
patches of the same size. Then each patch is approximately represented as linear combination of the dictionary atoms.
A HR patch $p_H$ could be reconstructed by the corresponding HR dictionary with the shared representation $\alpha$ obtained from LR patch $p_L$.

  \begin{equation} \label{eq2}
    p_L \approx \Upphi_1 \alpha,\ \ \ \  p_H = \Upphi_s \alpha
  \end{equation}

We denote $\Upphi_s \in \mathcal{R}^{m\cdot s^2 \times N}$ as the dictionary,
where $s$ is the upscaling factor. If the value of $s$
 is $1$, $\Upphi_s$ exactly becomes $\Upphi_1 \in \mathcal{R}^{m \times N}$ for the
 LR patches. $N$ is the size of the
whole dictionary, and $m = w \cdot w$ is the vectorization size of square patch with $w$ length.
Therefore, the size of $p_L$ is $m$,  while the size of $p_H$ for HR image is $m\cdot s^2$.

The SR task is ill-posed which means there are infinite $\mathbf{x}$ satisfying the Eq. (\ref{eq1}). The
sparse representation methods provide sparse prior for the representation of patch $p_L$ to regularize the task.
Sparse prior assumes $\|\alpha\|_0$, the coefficients of representation, should be as small as possible,
while still contains the low reconstruction error.
With the constraint above, SR task can be formulated as an optimization problem:
  \begin{equation} \label{eq3}
    \mathop{\min} \| \alpha \|_0  \ \ \ \ s.t. \ \ \| p_L -  \Upphi_1 \alpha \|_2^2 \leq \varepsilon
  \end{equation}

The aforementioned problem is still non-convex and NP-hard. However, work on \cite{donoho_2006}
shows that (\ref{eq3}) can be converted by replacing the objective with the $l^1$-norm:
  \begin{equation} \label{eq35}
    \mathop{\min} \| \alpha \|_1  \ \ \ \ s.t. \ \ \| p_L -  \Upphi_1 \alpha \|_2^2 \leq \varepsilon
  \end{equation}
Applied with Langrange multipliers, (\ref{eq35}) becomes a convex problem which can be efficiently solved by LASSO \cite{tibshirani_1996}.

  \begin{equation} \label{eq4}
    \alpha = \mathop{\arg} \mathop{\min}_{\alpha} \left\{  \| p_L - \Upphi_1 \alpha \|_2^2 + \lambda \| \alpha \|_1  \right\}
  \end{equation}
$\lambda$ denotes sparse regularization to balance the sparsity of the representation and the reconstruction error.
In conventional sparse representation method \cite{yang_2010}, we need external images to train
the dictionary $\Upphi_1$ and $\Upphi_s$. However, since it is not required in the proposed algorithm,
 the part of describing dictionary from training has been excluded.

\section{Hybrid Functions Sparse Representation} \label{3}

\subsection{Function-based dictionary}
  It is critically important to choose dictionary for sparse coding.
  Despite the training-based algorithm, another type of dictionary uses wavelets and curvelets
  which shares some similarity with proposed function-based dictionary.
  However, the ability of adapting different types of data for this dictionary is limited and do not
  perform well under sparse prior \cite{zhang_2015}.

  There are some advantages of using the function to generate the dictionary,
  in which we can leverage the property of the function to fine tune parameters or
  enlarge the dictionary to any scale as is required.
  The relation between dictionary and functions is connected as:
  \begin{equation} \label{eq5}
    \Upphi_1^k(i, j | \Theta^k) =  f^k (i, j, \Theta^k)
  \end{equation}
  Here, $\Theta=\{\Theta^1,\Theta^2,....,\Theta^N\}$ is the set of parameters for the whole dictionary.
   $\Upphi_1^k \in \mathcal{R}^m$ represents the $k^{th}$ element in the LR dictionary,
  which is the vectorized result from the patch with size $\mathcal{R}^{w \times w}$, so the integer pair $(i, j)$ maps
  the pixel value by coordinate from the raw patch to the $(i*w + j)^{th}$ position in $\Upphi_1^k$ ,
  $f^k$ and $\Theta^k$ is the $k^{th}$ function in dictionary and its corresponding parameter, respectively.
  \begin{equation} \label{eq6}
    f^k (i, j, \Theta^k) \in \left \{ \begin{array}{lll}
    \mathop{\arctan} \left(  (i \cdot \cos \theta^k + j \cdot \sin \theta^k + b^k) / \xi^k  \right) / \pi \\
     \sin (  (i\cdot \cos \theta^k + j\cdot \sin \theta^k + b^k) \cdot a^k  ) \\
     \cos \left(  \frac{\pi a^k}{w} (i + \frac{1}{2})  \right) \cdot \cos \left(  \frac{\pi b^k}{h} (j + \frac{1}{2})  \right) \end{array} \right.
  \end{equation}
  \begin{figure}
    \begin{tabular}{ccc}
      \includegraphics[width=0.33\linewidth]{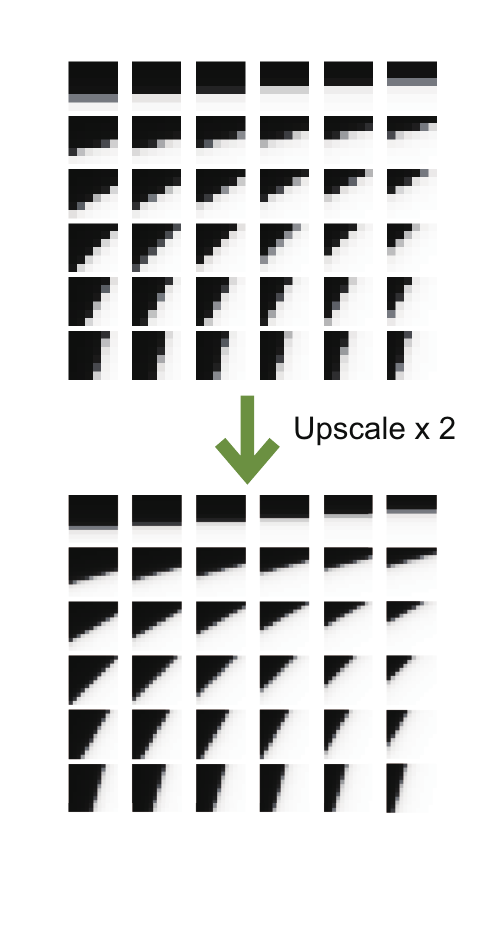} &s
      \includegraphics[width=0.33\linewidth]{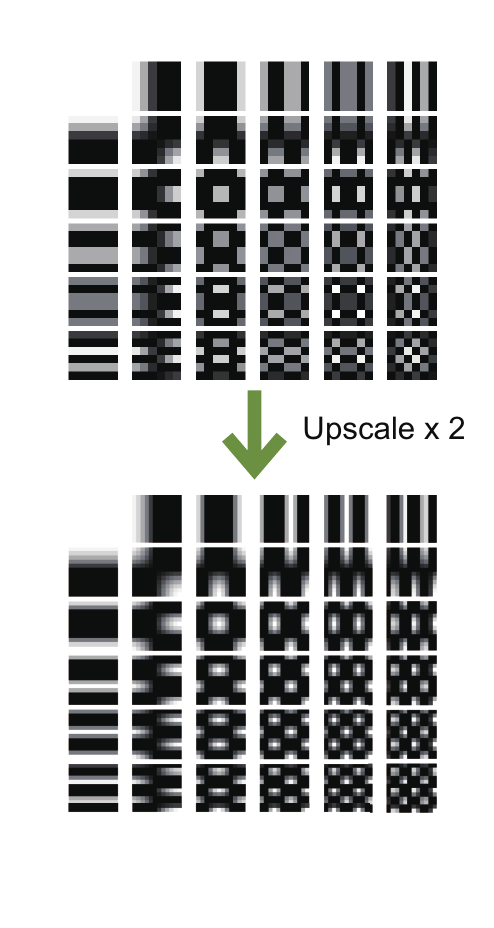} &
      \includegraphics[width=0.33\linewidth]{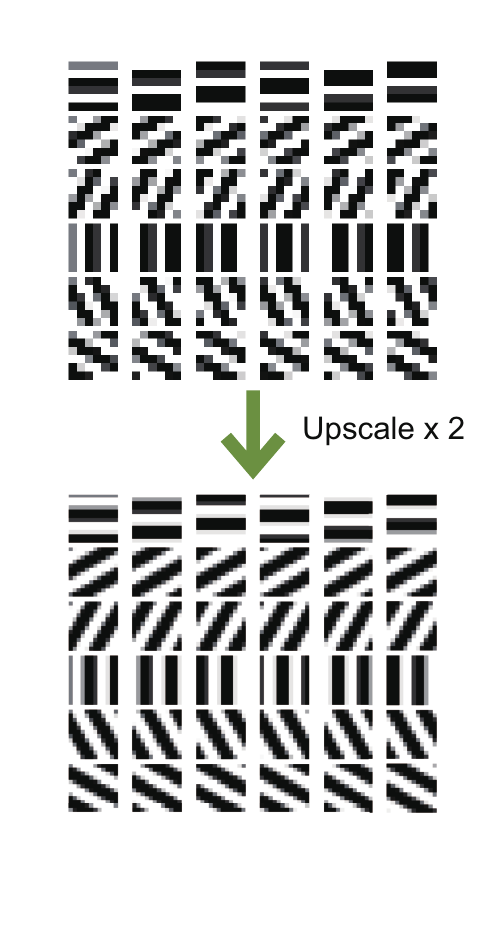} \\
      (a) & (b) & (c)
    \end{tabular}
    \caption{The visualization of LR patches and corresponding HR patches sampled from the hybrid of
    (a) AHF, (b) DCT function and (c) Sin function.
    The first row are the LR patches from $\Upphi_1$, the second row are
    the HR patches from $\Upphi_2$. }
    \label{fig01}
  \end{figure}

  As is shown in Eq. (\ref{eq6}), three categories of function are combined here.
  The approximate Heaviside function can be used to fit the edge in patch, where one-dimensional AHF
  is $\psi(x) = \frac{1}{2} + \frac{1}{\pi} \arctan(\frac{x}{\xi})$, and the parameters $\xi$ control the
  smoothness of margin between $1$ and $0$. Then the function is extended to two-dimensional case for our task,
  thus
  $f_{AHF}(\mathbf{z}) = \psi(\mathbf{z} \cdot \mathbf{d}^T + b)$, where $\mathbf{d} = (\cos \theta , \sin \theta)$
  and $\mathbf{z} = (i, j)$ is the coordinate in patch.
  Since sine is a simple periodic function and is suitable for representing stripes,
  we adopt it as one of the basic functions
  and extend it to two-dimensional case as what we did for AHF.
  $f_{Sin}(\mathbf{z}) = \sin (a \mathbf{z} \cdot \mathbf{d}^T + b)$,
   where parameter $a$ and $\theta$ for 2D-sine function controls the intensity and direction of the stripes,
   respectively.
  The idea of using over-complete discrete cosine transformation (DCT) as one of the functions is inspired by \cite{ksvd_2006},
  the pixel value in DCT patch is generated by $\cos \left(  \frac{\pi a}{w} (i + \frac{1}{2})  \right) \cdot
   \cos \left(  \frac{\pi b}{h} (j + \frac{1}{2})  \right)$.
   We have designed several functions and tried various mixed strategies, then we find the hybrid of
   these three functions work better than single functions or other mixed forms.

  \begin{equation} \label{eq7}
    \Theta^k \in \left \{ \begin{array}{lll} \{ \theta_k, b_k, \xi_k \} \\
    \{ \theta_k, a_k, b_k  \} \\ \{  a_k, b_k  \}  \end{array} \right.
  \end{equation}

  The parameter selections of three functions are listed in the Eq. (\ref{eq7}).
  Fig. \ref{fig01} depict upscaling process using three functions above, respectively.
  One of the advantages of HFSR approach is that the dictionary for
  reconstructing the HR image with a specific size could be directly obtained by functions.
  Generating the HR dictionary and LR Dictionary can be connected in a similar way.
  \begin{equation}
    \Upphi_s(i, j | \Theta) = \Upphi_1(i/s, j/s | \Theta) = f (i/s, j/s, \Theta)
  \end{equation}

\subsection{Multi-scale refinement}

The patches are sampled in raster-scan order from the raw LR image
and processed independently.
 The full representation for each patch is computed by two processes.
 Sparse coding is firstly applied to get coarse representation $\alpha_{coarse}$,
 and then the proposed multi-scale refinement is used to generate fine representation $\alpha_{fine}$.

Suppose $\alpha$ is the optimal solution,
then the difference of $p_L - D \Upphi_s \alpha$
should equal to zero, where $D$ is the downsampling operator.
However, the condition is typically unsatisfied with $\alpha_{coarse}$,
and the difference usually contains some residual edges.
To address this issue, we pick $p_L - D \Upphi_s \alpha$ as a new
LR patch to fit. Then recompute a representation $\Delta \alpha$ for the new residual input
 and add $\Delta \alpha$ to $\alpha$ for updating.
 When the values in residual image are small enough,
 refinement ends with a better representation $\alpha_{fine}$.

 Generic refinement method downsample the patch from HR patch with specific upscaling factor $s$.
To utilize the scalable property of our function-based dictionary, we also put
forward a new refinement algorithm called multi-scale refinement.
Scaling up an image with $s$ by interpolation-based algorithm can be
viewed as a process of iteratively expanding an image by multi-steps.
During each step, images are enlarged slightly compared with the ones in the previous step.
As such, our proposed algorithm could perform this procedure due to the scalable
property.
 Here, it is necessry to make an assumption that during the iterative process of
 expanding an image,
 if representation for each patch is well-obtained,
 the reconstruction could be of high quality at each zooming step
 rather than perform well only in a certain scale but distort at other optional
 scaling factors.
 This assumption can be viewed as a regularization that saves the ill-posed problem
 from massive solutions.
Therefore, we make the refinement for each
scale to enforce the stability of its reconstruction results.
 The difference of $k^{th}$ scale refinement with scaling factor $s_k$ hence
 becomes $p_L - D \Upphi_{s_k} \alpha$.
Experiments demonstrate that multi-scale
refinement improved the results and is less time consuming.

\begin{algorithm}[H]
\DontPrintSemicolon

  \KwInput{ \{$p_L$ : low resolution patch \}   \{$\Upphi$ : dictionary\} }
  \KwOutput{ \{$\alpha_{fine}$: representation\} }
  $\alpha_{coarse} = \mathop{\arg} \mathop{\min}_{\alpha} \| p_L - \Upphi_1 \alpha \|_2^2 + \lambda_1 \| \alpha \|_1 $ \\
  $k = 1$ \\
  $p_L^0 = p_L$\\
  $\alpha^0 = \alpha_{coarse}$ \\
  \For{$\bar{s}\in scales$}
    {
      \For{$iter$ = 1:\ $iters_{\bar{s}}$}
      {
        $p_L^k = p_L^{k-1} - D \Upphi_{\bar{s}} \alpha^{k-1}$ \\
        $\alpha^{k} = \mathop{\arg} \mathop{\min}_{\alpha} \| p_L^k - \Upphi_{1} \alpha \|_2^2 + \lambda_2 \| \alpha \|_1 $ \\
        $k = k + 1$ \\

      }
    }
  $\alpha_{fine} = \alpha_{coarse} + \sum_{i=1}^k \alpha^i$

\caption{HFSR with multi-scale refinement for patch approximation}
\end{algorithm}

\subsection{Image reconstruction from patches}

\begin{figure}
  \begin{tabular}{c}
    \includegraphics[width=1.0\linewidth]{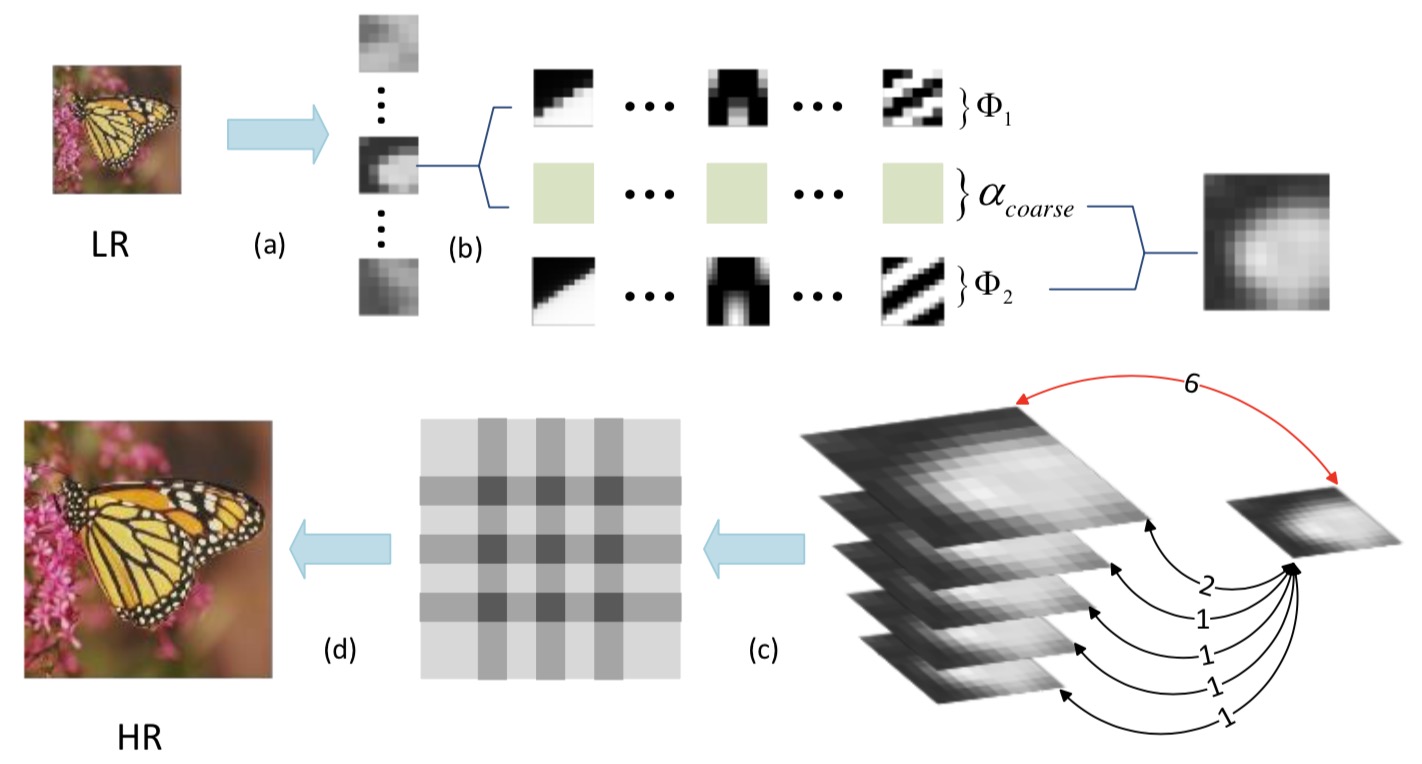}
  \end{tabular}
  \caption{Overview of the proposed HFSR method. Process (a) sample the patches from the LR image, (b) compute
  the coarse representation for each patch, (c) apply multi-scale refinement to obtain fine representation,
   (d) reconstruct the HR image from overlaped patches. The red arrow represents the conventional
   refinement used for comparison.
  }
  \label{fig2}
\end{figure}
The whole image can be reconstructed after calculating the representation for each patch in raster-scan order.
 However the borders of adjacent patches are not compatible which may cause patchy result.
To deal with this issue, the adjacent patches are sampled with overlaps
which would assign multiple values to each pixel.
General method is to average these pixel values for reconstruction.
Based on the aforementioned analysis, objective loss represents the reconstruction quality of the patch
 and can be involved in reconstruction process for more precise result.
 Following this idea, we make weighted summation according to the reconstruction
 objective loss to recover the pixel value.
\begin{equation}
  HR(x,y) = \frac{1}{\sum_{j=1}^{|p|}loss^j} \sum_i^{|p|} loss^i \cdot p_H^i(x^i,y^i)
\end{equation}
$HR(x,y)$ represents the pixel in HR image with coordinate $(x,y)$. Suppose the pixel is
covered by $|p|$ patches, and $p_H^i(x^i,y^i)$ is the corresponding value in $i^{th}$ constructed
patch while $loss^i$ is the objective loss.
 The strategy works better than generic average approach especially when the objective loss
 vary dramatically among different patches.
 Fig. \ref{fig2} demonstrates the concrete procedures of HFSR algorithm.

\section{Experimental Results} \label{4}

In this section, the proposed algorithm HFSR is evaluated with two classical interpolation methods
and one of the state-of-the art methods proposed by glasner et al. \cite{glasner_2009}.
The hyperparameter and some implementation details are also analyzed and discussed here.
ScSR \cite{yang_2010} is not compared here because it requires
HR images for training.
AHF \cite{deng_2016} approach is not sparse coding method
and needs the inverse of matrix to solve ADMM problem which runs extremely slowly,
the comparison is hence excluded either.
All experiments are implemented in Python 3.6 on Macbook Pro
of 8Gb RAM and 2.3 GHz Intel Core i5.

The performance is measured by peak signal-to-noise ratio (PSNR),
and upscaling factor $s$ is set to $2$ for all experiments.
RGB input images are first converted into $YC_BC_R$ color space with
three channels. $Y$ channel represents luma component which is
processed in most SR algorithms.
Therefore, we upscale it independently by the proposed model.
To evaluate the result, we only compared the $Y$ channel by PSNR.
The remaining two channels $C_B$ and $C_R$ are enlarged by bicubic
interpolation so that $YC_BC_R$ image can be transformed
to the original RGB image for visualization.
Grayscale input images can be directly applied by the proposed method and evaluation.

\newcommand\bg[1]{\textcolor{brown}{\textbf{#1}}}

\begin{table}[t] 
  \caption{The PSNR (dB) results obtained through the experiments, HFSR is the proposed method
  with conventional refinement, while HFSR(multi-scale) is the proposed method with multi-scale Refinement. 
  The bold means the highest score among the non-learning method while the brown color represents the second best score.}
\begin{tabular*}{\textwidth}{|l| @{\extracolsep{\fill}}ccccccc |}
\hline
                      & \multicolumn{1}{c|}{baboon} & \multicolumn{1}{c|}{barbara} & \multicolumn{1}{c|}{Bridge} & \multicolumn{1}{c|}{Coastguard} & \multicolumn{1}{c|}{Comic} & \multicolumn{1}{c|}{Face} & Flowers \\ \hline
                        \multicolumn{8}{|l|}{Non-Learning method} \\ \hline
 nearest neighbor     &          24.21        &       27.18           &        27.94          &    28.19              &         24.61        &     33.63             & 28.41          \\ \cline{1-1}
     bicubic          &         24.67         &       27.94           &        28.96          &      29.13            &         26.05        &        34.88          & 30.43          \\ \cline{1-1}
     glasner          &         25.11         &    \textbf{ 28.54 }   &    \textbf{29.66 }    &   \textbf{29.80 }     &         26.66        &  \textbf{35.24 }      & \textbf{31.48} \\ \cline{1-1}
     HFSR             &  \bg{25.12}           &       28.15           &        29.50          &        29.47          &         \bg{26.88}   &        34.45          & 31.27          \\ \cline{1-1}
    HFSR(multi-scale) &       \textbf{25.13}  & \bg{28.16}            &      \bg{29.52 }      &        \bg{29.49}     &   \textbf{26.90}     &     \bg{34.46}        & \bg{31.29}     \\ \hline 
    \multicolumn{8}{|l|}{Learning-based method} \\ \hline
    ScSR              &    25.24              &       28.52           &       29.97           &      30.28            &         27.66        &      35.56            & 32.58          \\ [0.5ex] \hline \hline
                      & \multicolumn{1}{c|}{Foreman} & \multicolumn{1}{c|}{Lenna} & \multicolumn{1}{c|}{Man} & \multicolumn{1}{c|}{butterfly} & \multicolumn{1}{c|}{Pepper} & \multicolumn{1}{c|}{PPT3} & Zebra \\ \hline
                      \multicolumn{8}{|l|}{Non-Learning method} \\ \hline
 nearest neighbor     &       30.35           &          32.35        &       28.01           &       30.19           &        31.09          &       25.05           & 27.37 \\ \cline{1-1}
      bicubic         &       32.66           &          34.74        &       29.27           &       32.97           &        33.08          &       26.85           & 30.68 \\ \cline{1-1}
      glasner         &  \textbf{34.15 }      &     \textbf{35.78}    &  \textbf{30.32}       & \textbf{ 36.22 }      &  \textbf{35.08  }     & \textbf{ 29.65}       & 31.13 \\ \cline{1-1}
      HFSR            &       33.56           &          35.07        &       29.72           &       33.98           &        33.67          &       27.46           & \bg{31.56}     \\ \cline{1-1}
    HFSR(multi-scale) &     \bg{33.57 }       &        \bg{35.10}     &    \bg{29.74}         &      \bg{34.00 }      &     \bg{33.68}        &     \bg{27.47}        & \textbf{31.59} \\ \hline
    \multicolumn{8}{|l|}{Learning-based method} \\ \hline
    ScSR              &    34.45              &       36.21           &       30.46           &      35.92            &         34.12         &      28.97            & 32.99          \\ \hline
\end{tabular*}
\label{tab1}
\end{table}

The priority is to firstly tune the parameters in dictionary. For
AHF, $\theta$ is sampled from $16$ angles,
which are distributed evenly on $[0, 2\pi]$.
$b$ is sampled from $[-6, 6]$ with $12$ isometric intervals. $\xi$ is sampled from $[0.1, 10^{-4}]$.
For Sin function , $6$ angles are ranged form $[0, \pi]$ for $\theta$,
$b$ is sampled from $[0, 1, 2, 3, 4, 5, 6]$ and $a$ is sampled from $[2.5, 2.25, 2]$.
For DCT function, $a$ is average sampled from $[0, 5]$ with $1$ distance. $b$ are sampled in the same way as
$a$.
Then the dictionary is composed by parameter combinations and excludes
 those whose norm of its corresponding constructed patch less than $1.0$.
 The total size of the dictionary in our experiment is $334$.

The size of patches sampled from the LR image is $6 \times 6$
and the overlap of the patch is set as $4$.
Smaller size of the patch like $5\times5$
may work better when the image details are more complex but it is more time consuming.
In original refinement, all of the $6$ iterations are operated on the patch
with fixed upscaling factor $2$.
In multi-scale refinement, patches are enlarged from size $6\times 6$ to $12\times 12$ where
the upscaling factors are sampled
from $[\frac{7}{6},\frac{8}{6},\frac{9}{6},\frac{10}{6},\frac{11}{6},\frac{12}{6}]$,
the number of iterations $iters_{\bar{s}}$ in each scale
 with corresponding upscaling factor, in order, are $[1,1,1,1,0,2]$.

Large sparse regularization parameters $\lambda$ will degrade the quality
for $\alpha$ to approximate $p_L$
but lead better upscaling results for what it represents. Small 
$\lambda$ usually
gets less reconstruct error for $p_L$,  but it is easy to lose
the fidelity when recovering the HR image.
Hence, it is important to select an optimal value $\lambda$
so as to balance the two effects above.
Ultimately, the sparse regularization $\lambda_1$ for coarse sparse representation and $\lambda_2$ for refinement
are both tuned as $10^{-4}$.

%

\begin{figure}[t] 
    \begin{tabular}{ccccc}
      \includegraphics[width=0.2\linewidth]{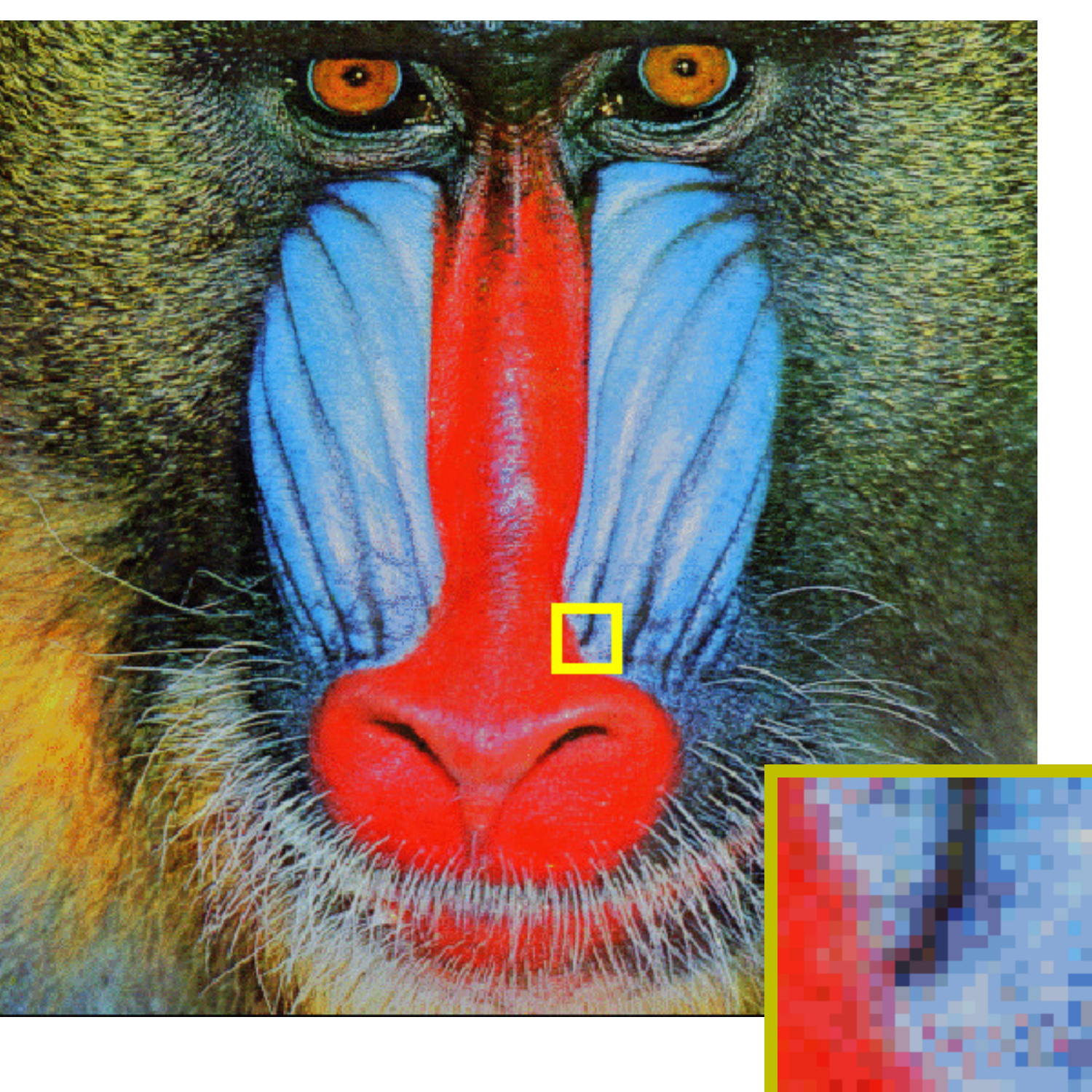} &
      \includegraphics[width=0.2\linewidth]{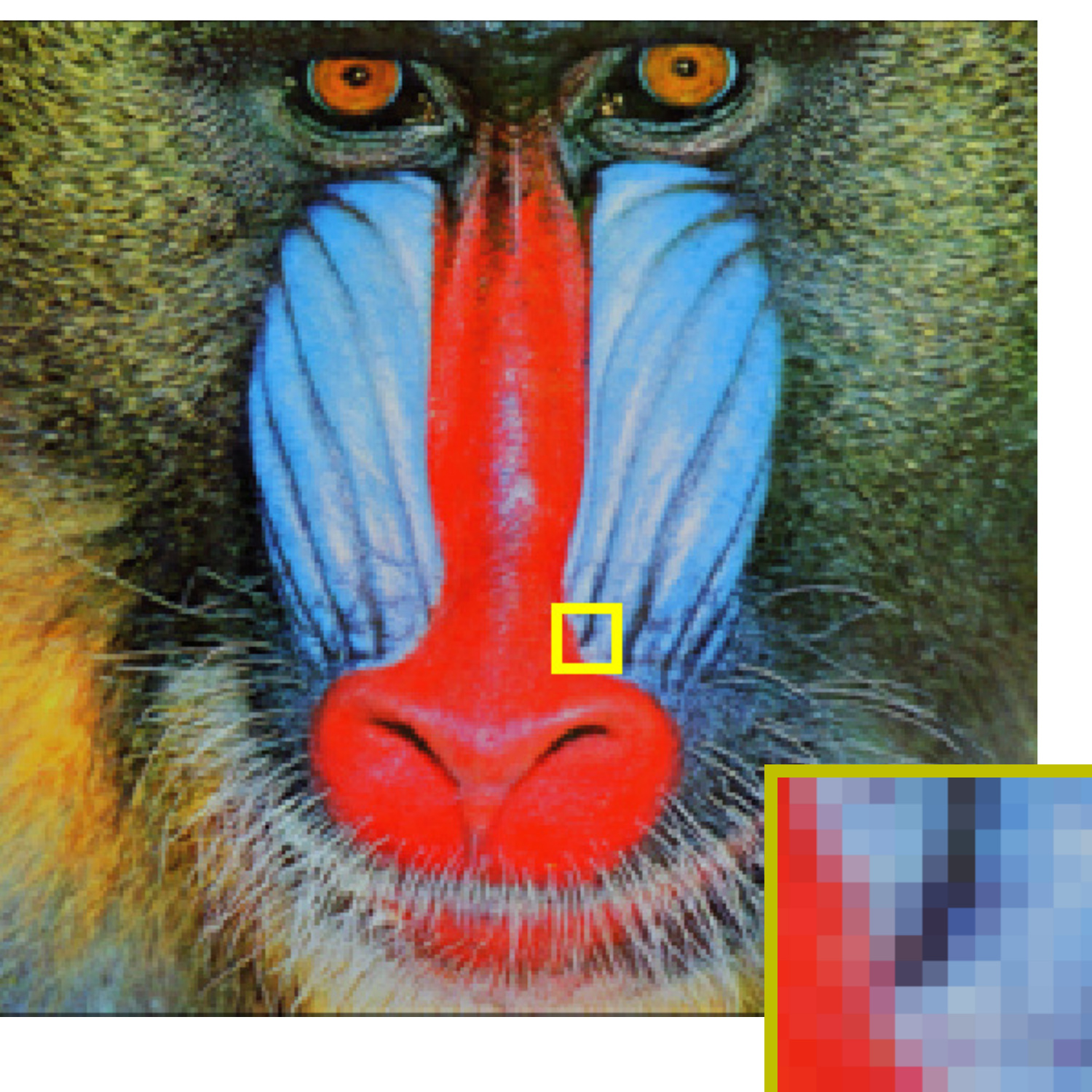} &
      \includegraphics[width=0.2\linewidth]{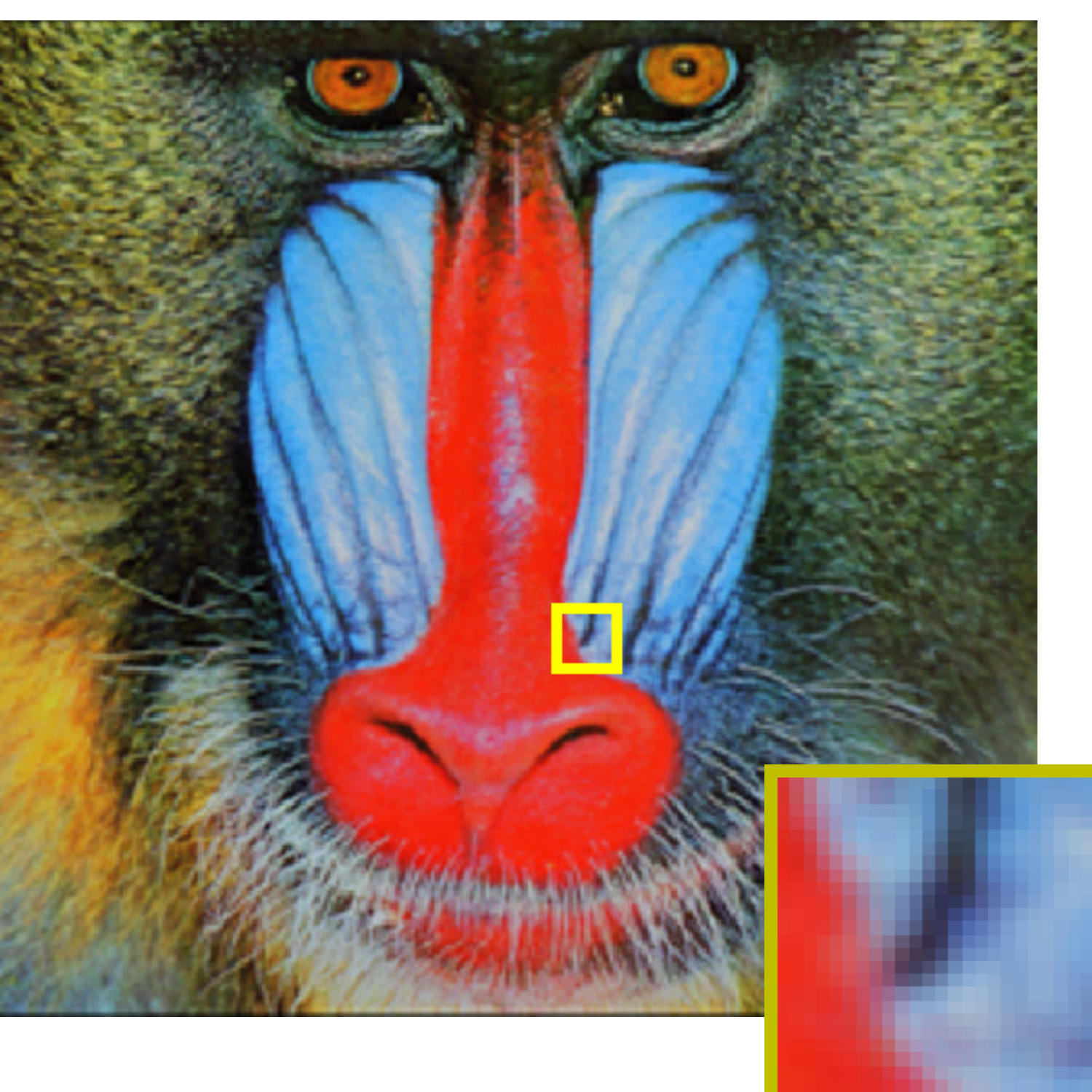} &
      \includegraphics[width=0.2\linewidth]{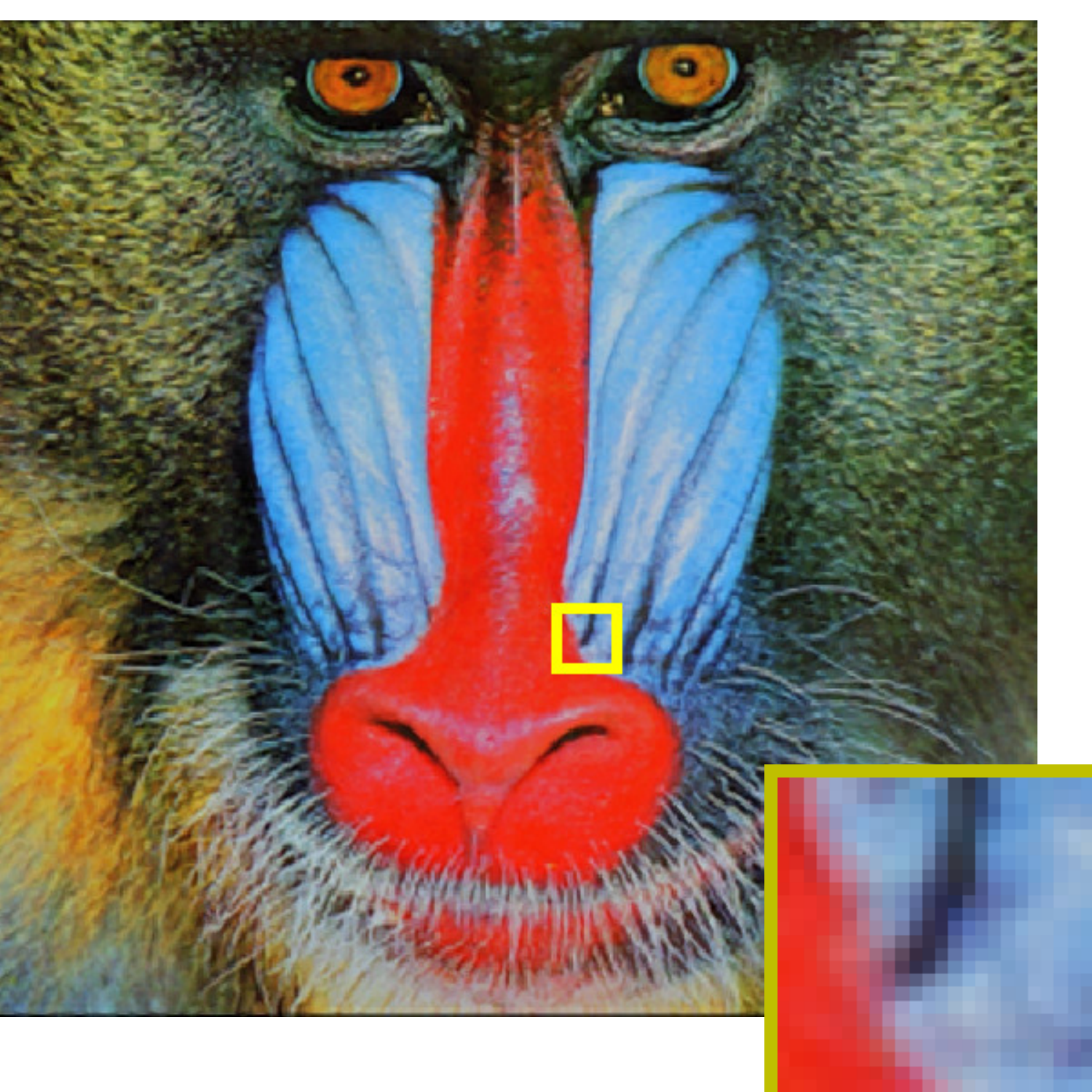} &
      \includegraphics[width=0.2\linewidth]{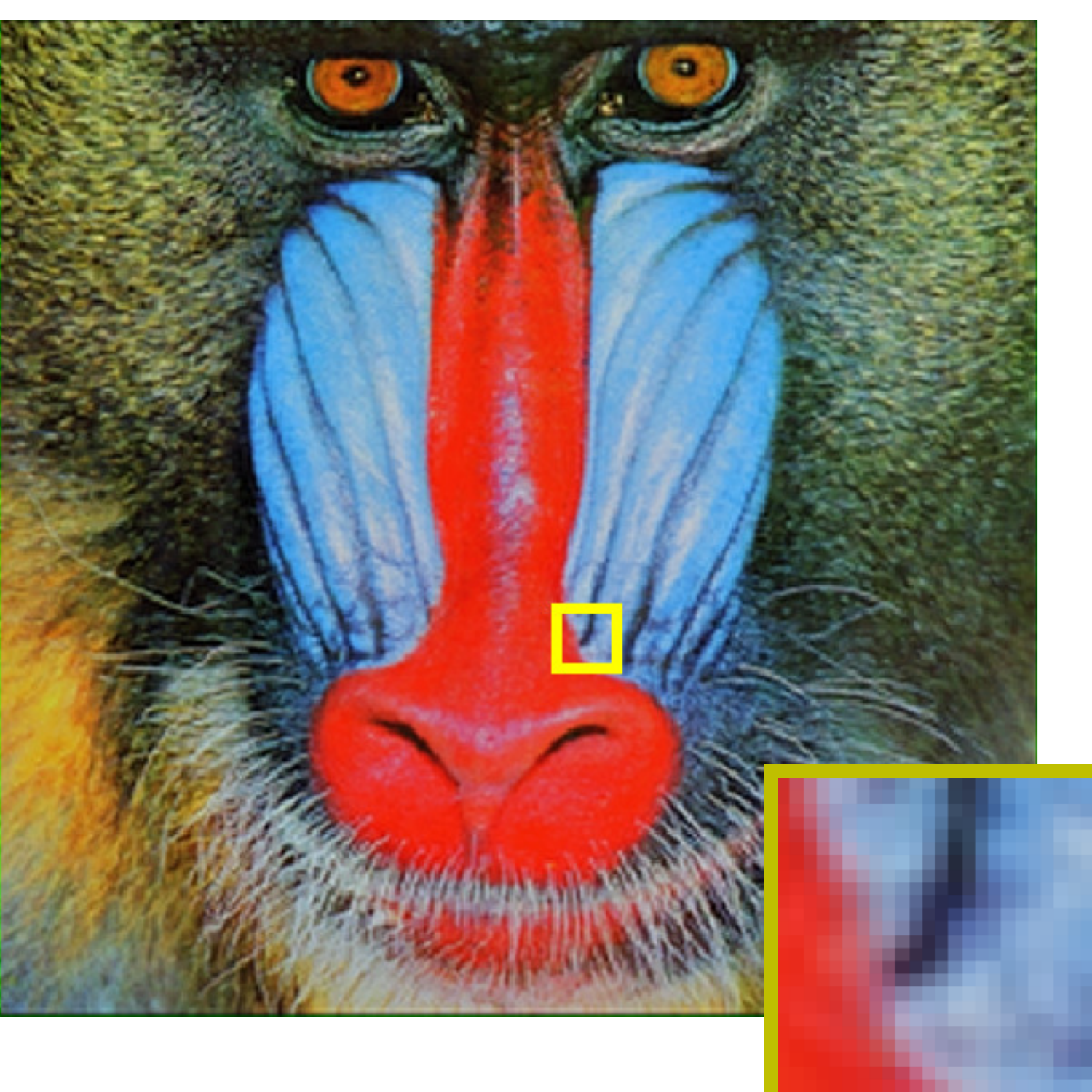} \\
      \includegraphics[width=0.2\linewidth]{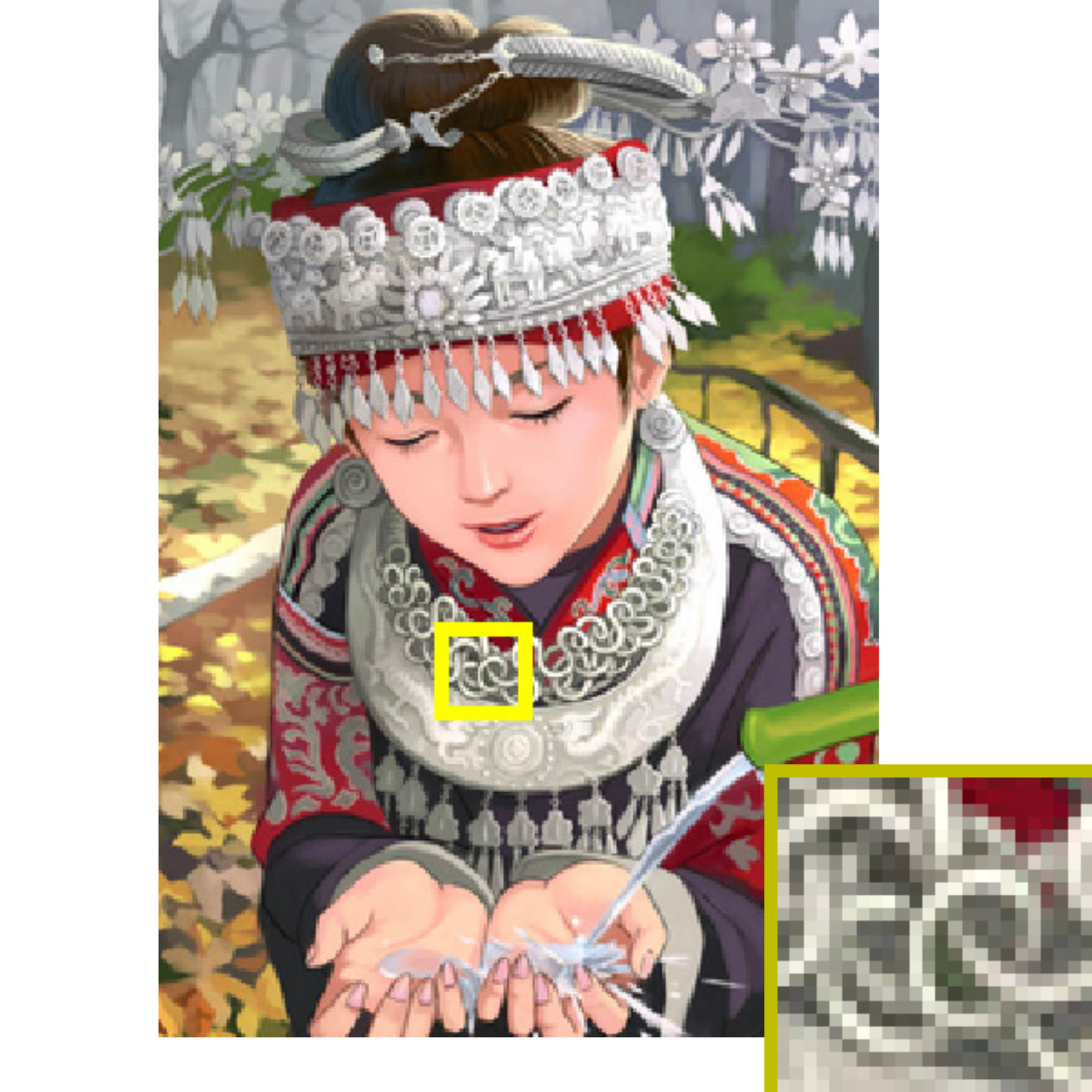} &
      \includegraphics[width=0.2\linewidth]{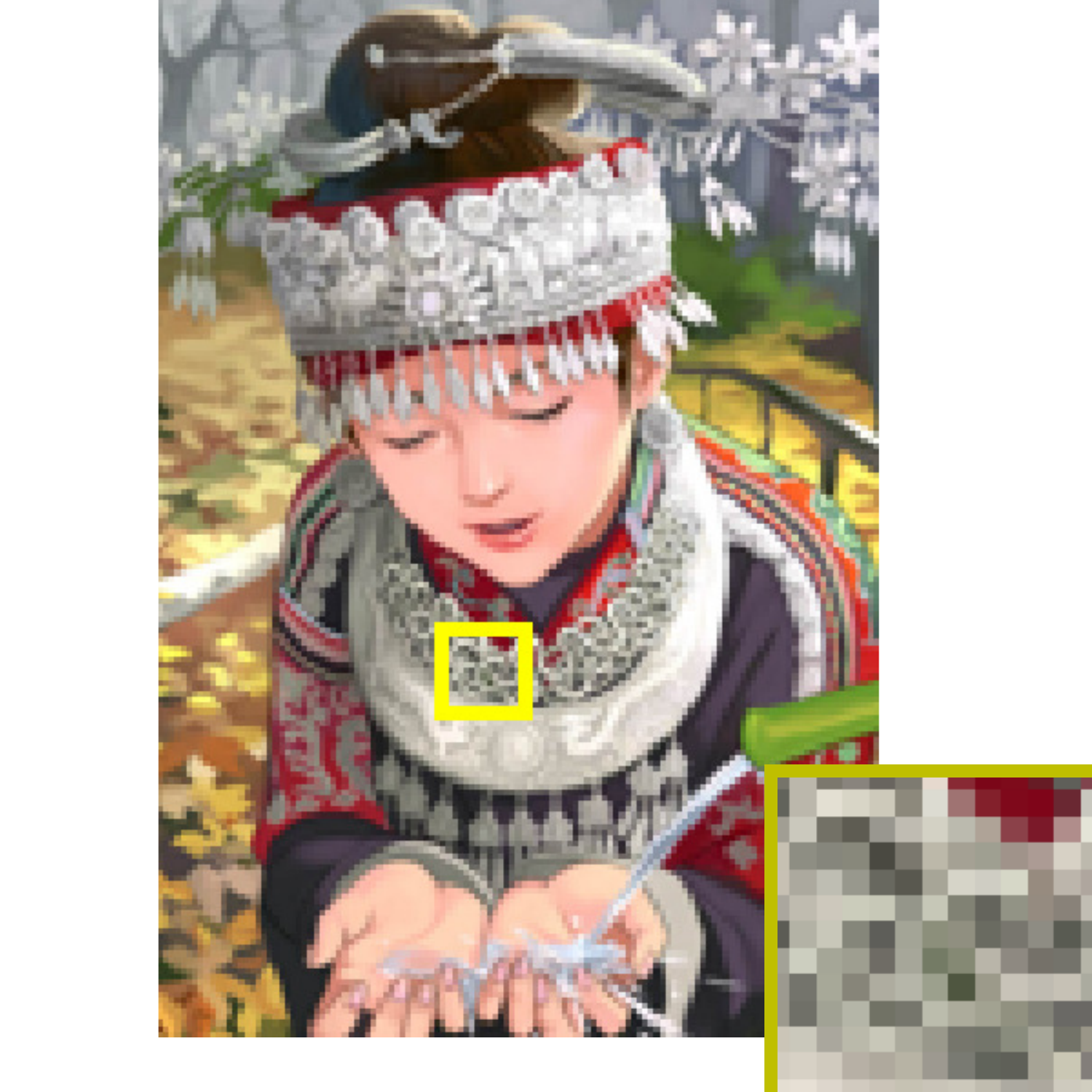} &
      \includegraphics[width=0.2\linewidth]{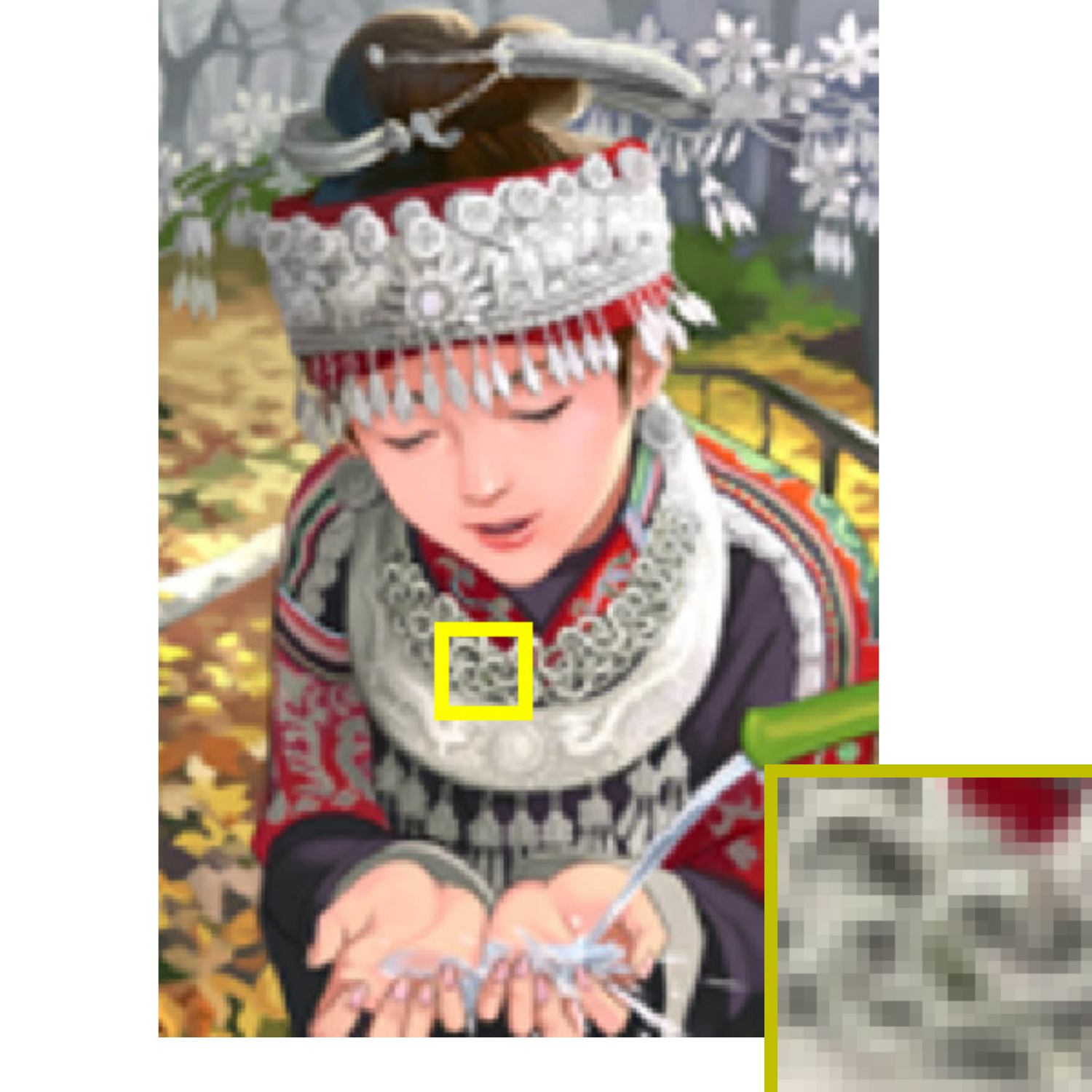} &
      \includegraphics[width=0.2\linewidth]{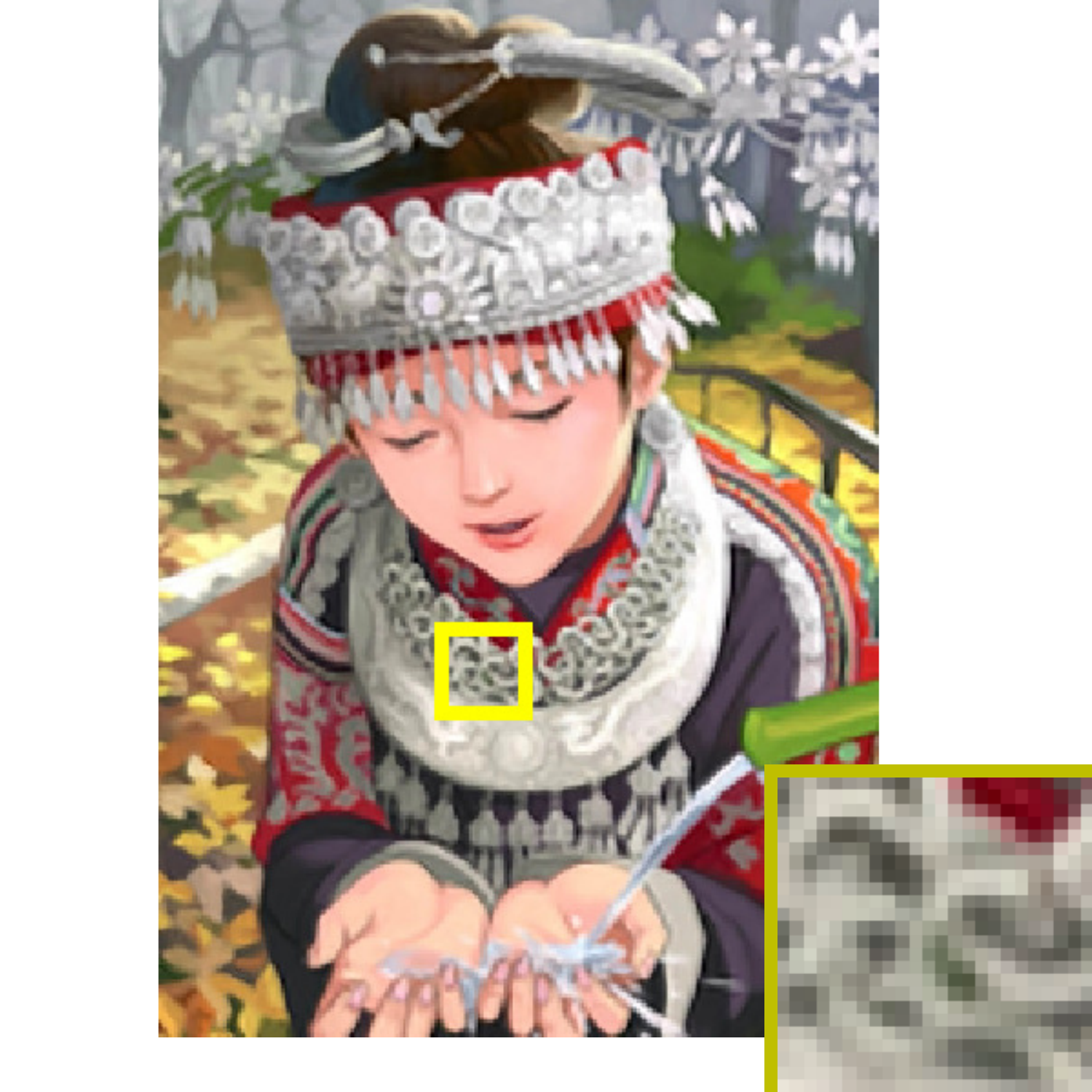} &
      \includegraphics[width=0.2\linewidth]{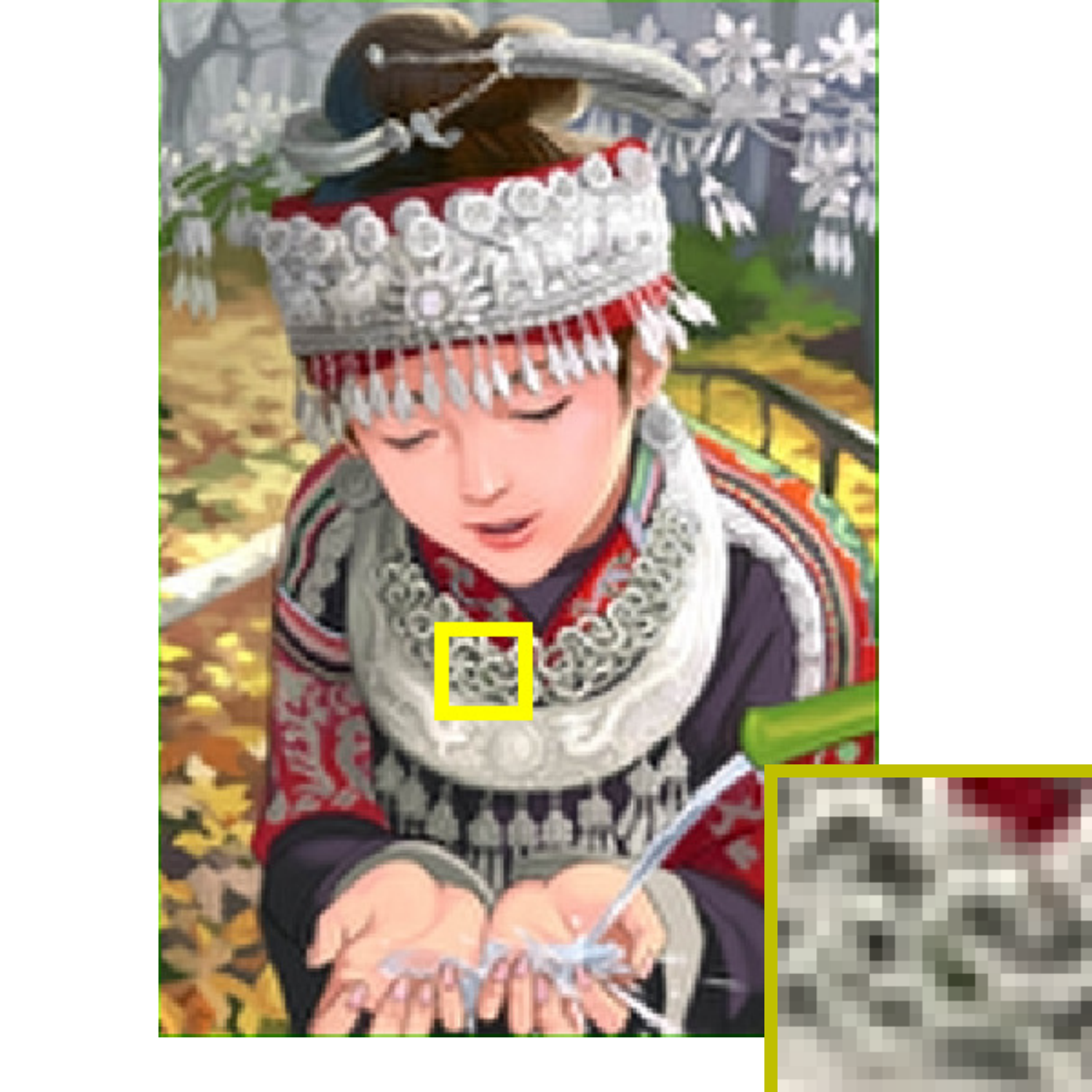} \\
      \includegraphics[width=0.2\linewidth]{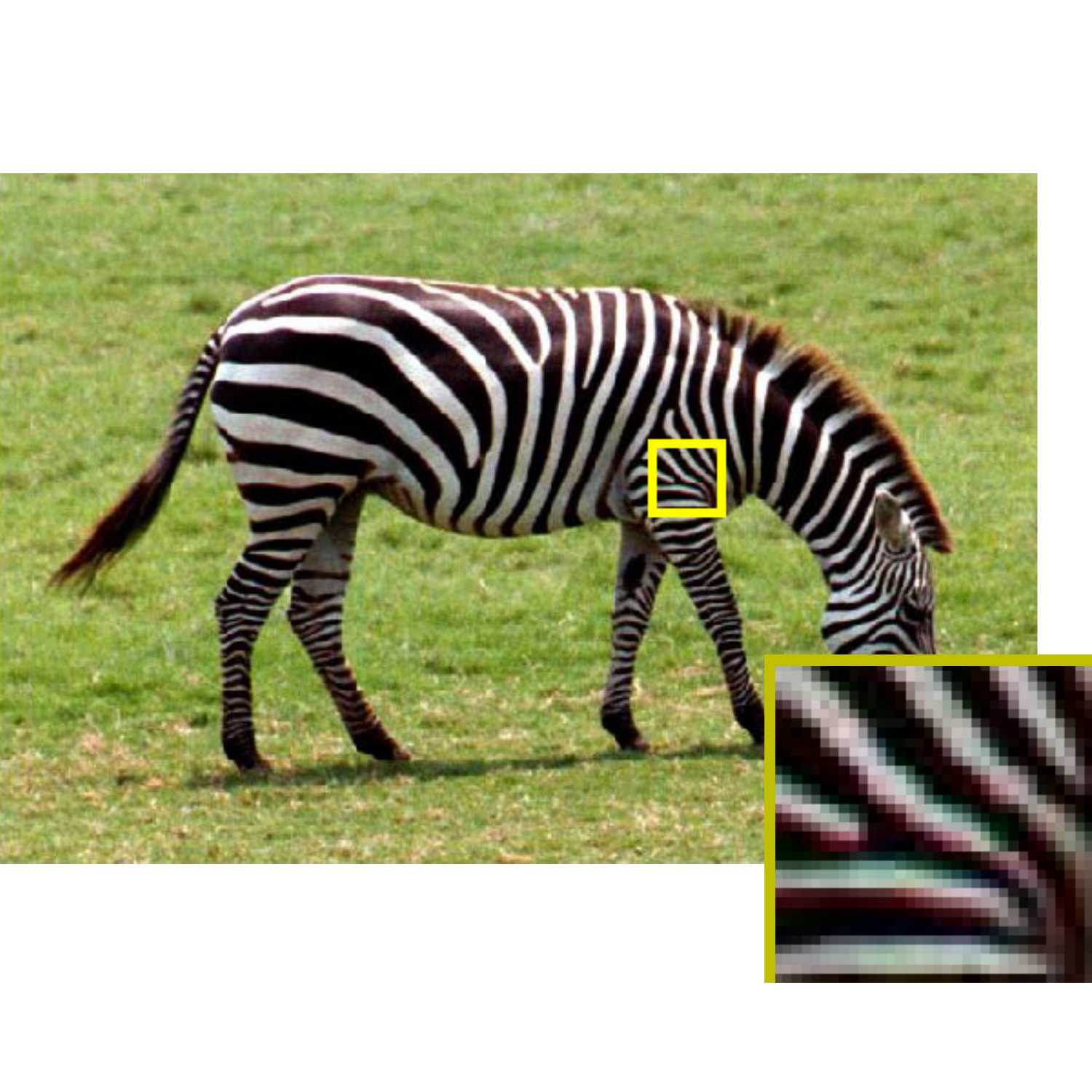} &
      \includegraphics[width=0.2\linewidth]{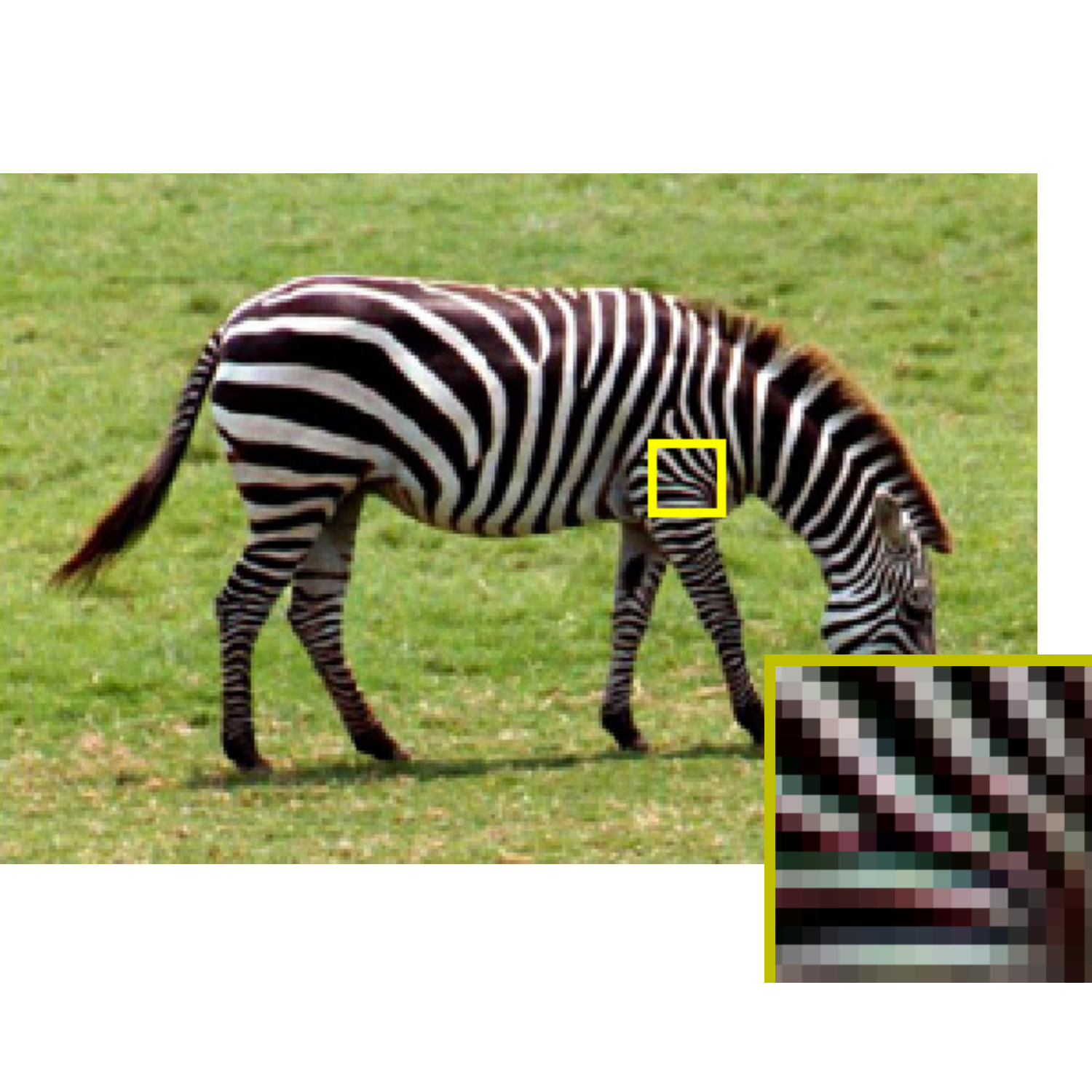} &
      \includegraphics[width=0.2\linewidth]{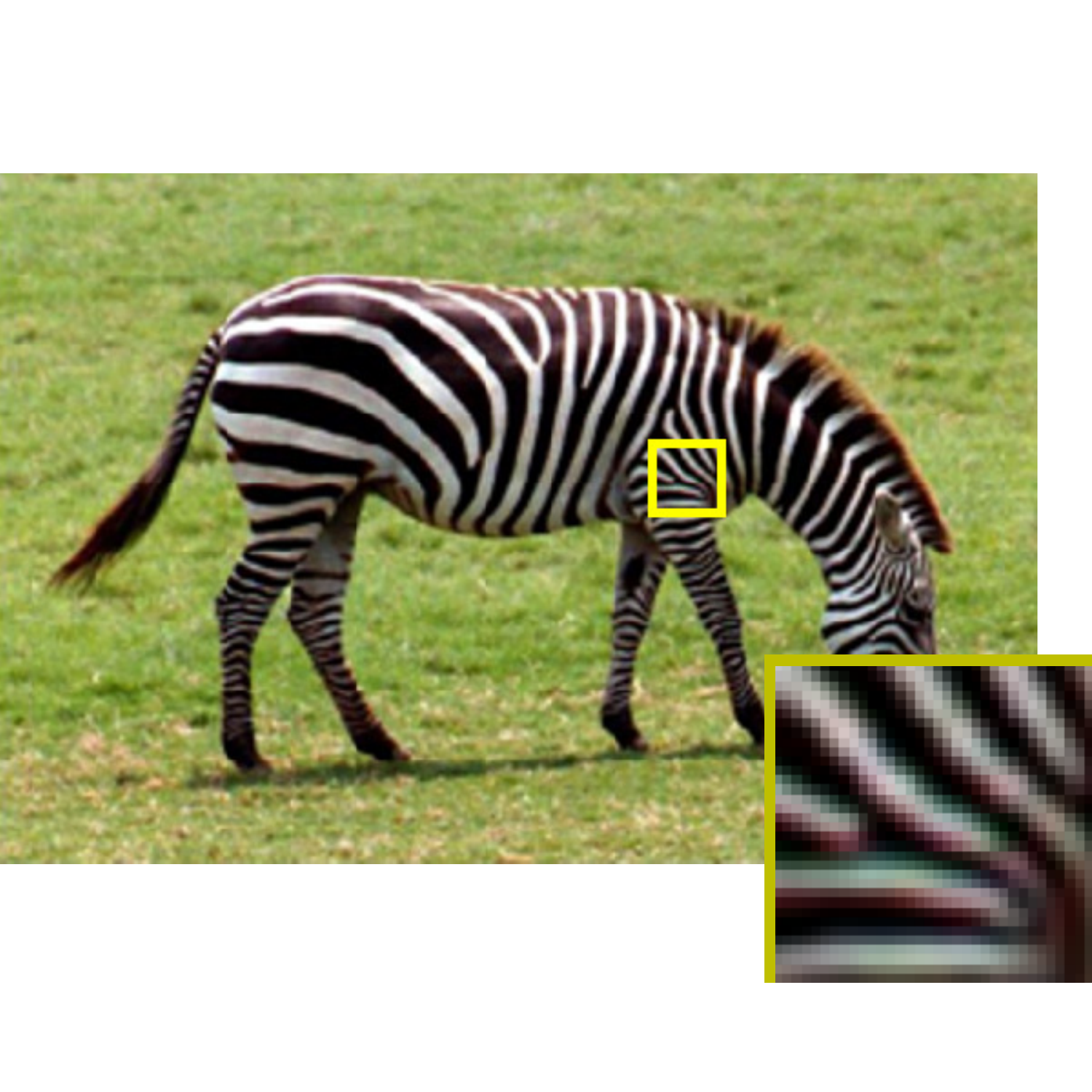} &
      \includegraphics[width=0.2\linewidth]{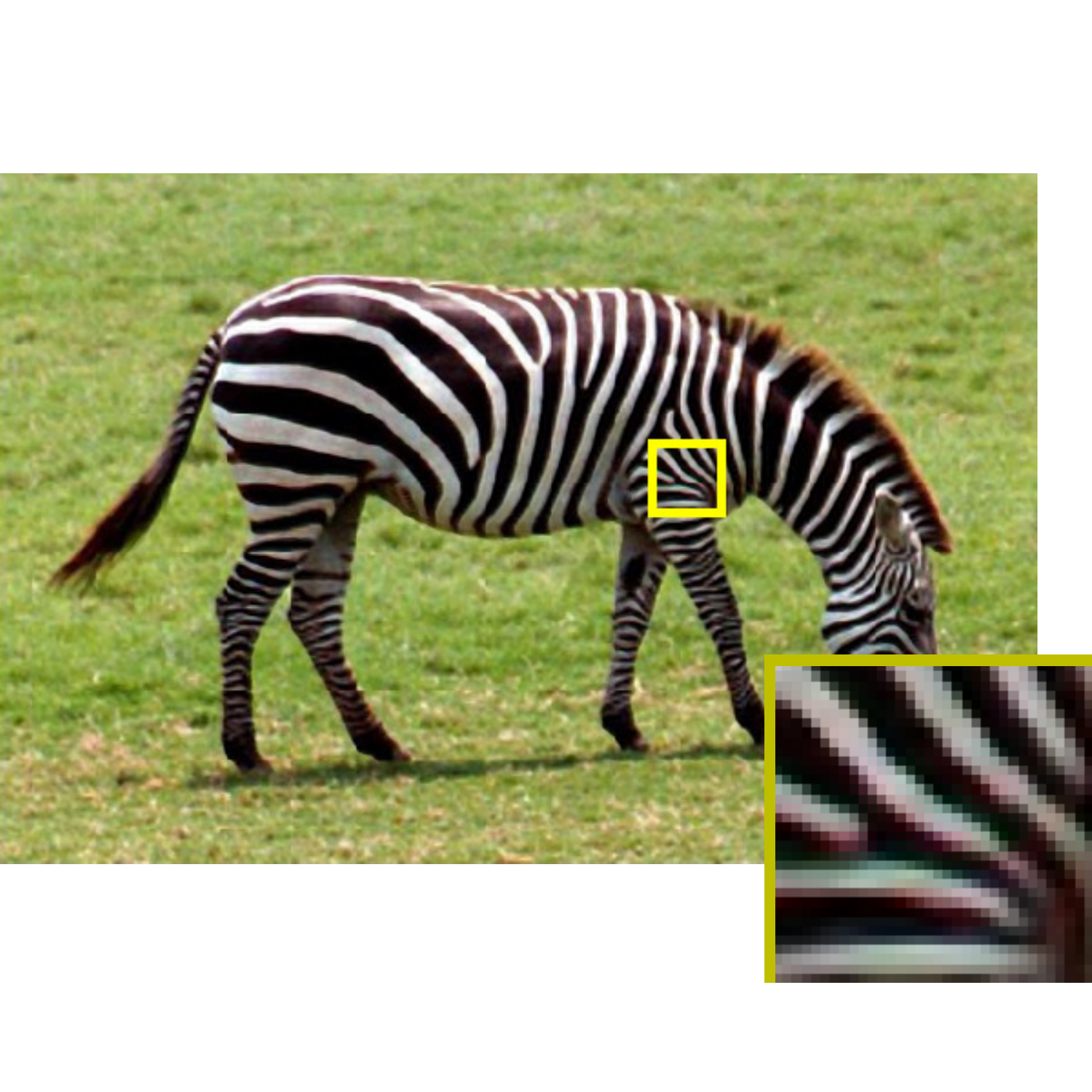} &
      \includegraphics[width=0.2\linewidth]{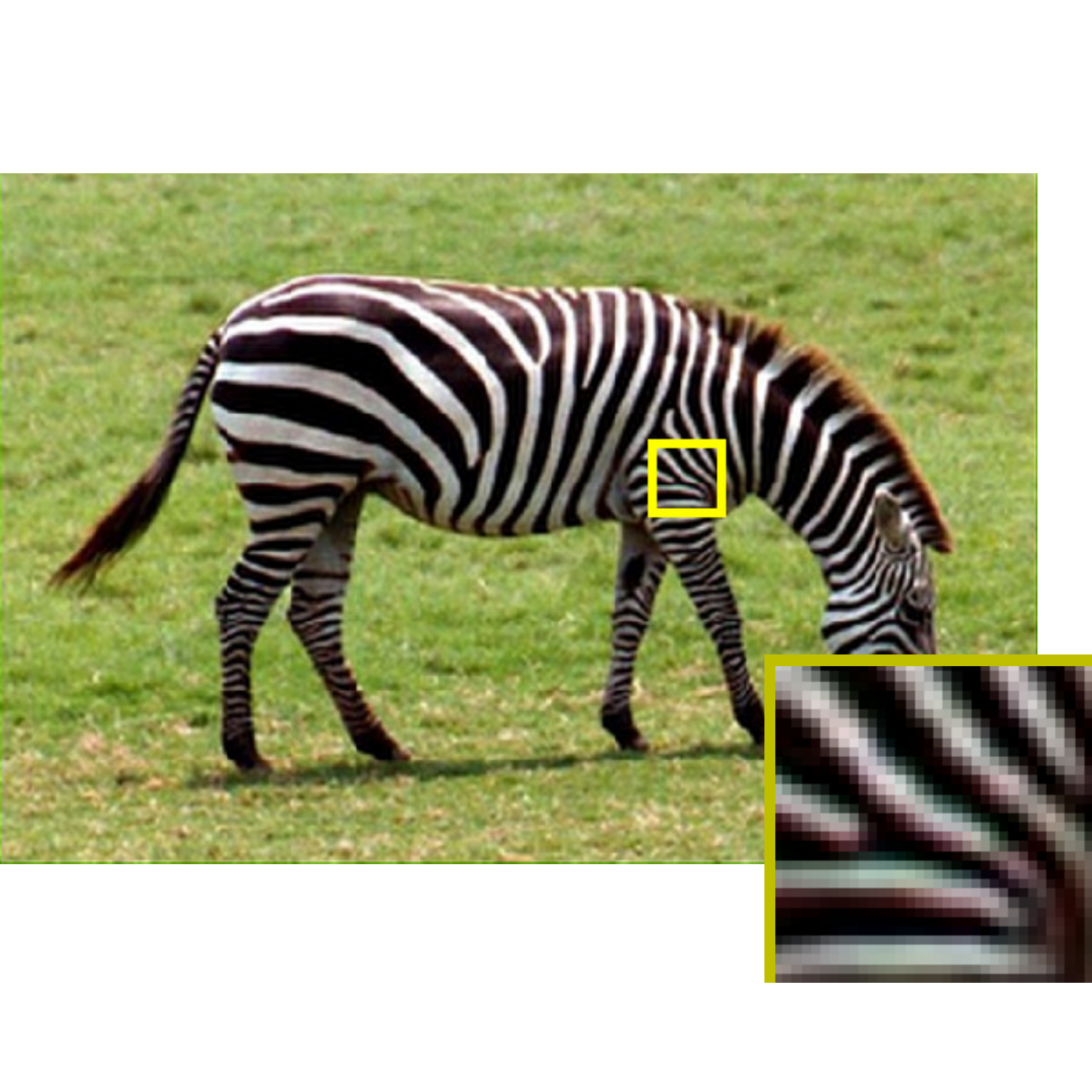} \\
      (a) HR & (b) NN & (c) Bicubic & (d) Glasner & (e) HFSR
    \end{tabular}
    \caption{ Results of images including baboon, comic and zebra with upscaling factor 2 (from top to bottom)
      and comparison of HR(Ground truth), NN, Bicubic, Glasner\cite{glasner_2009} and proposed HFSR (from left to right) }
    \label{fig3}
  \end{figure}

The experimental results from Table \ref{tab1} on the dataset Set14 show that HFSR
performs much better than conventional interpolation-based methods,
and multi-scale refinement is verified to improve PSNR score.
Furthermore, in Fig. \ref{fig3}, we select three pictures for visualization 
in which HFSR outperforms the glanser \cite{glasner_2009} and find that all of them 
contain rich textures. This is probably because our dictionary well contains the 
corresponding features and therefore can recover the patch robustly.

\section{Conclusion} \label{5}

This paper proposed a framework of
using function-based dictionary for sparse representation in SR task.
In our HFSR model, AHF, DCT function and sine function are combined
for patch approximation to form a hybrid function dictionary.
The dictionary is scalable without additional training,
and this property is utilized to design the multi-scale
refinement to improve the proposed algorithm.

The experiment is performed on the Set14 benchmark, and results show that the HFSR algorithm
performs well on a certain type of images which contains complex details and contexts.
For future improvements, the dictionary comprising more functions can be explored.
Many interesting topics are remained, including using the learning-based method
to fine tune parameters of the dictionary or apply HFSR to other domain such as
image compression or image denoising. To encourage future works of proposed algorithm
and discover effects in other applications, we public all the source code
and materials on website: https://github.com/Eulring/Hybrid-Function-Sparse-Representation.

{
\bibliographystyle{splncs04}
\bibliography{main.bbl}
}

%
%
%

\end{document}